\pdfoutput=1

\documentclass[11pt, a4paper]{article}

\usepackage{authblk}
\usepackage{acl}
\usepackage{times}
\usepackage{latexsym}

\usepackage[T1]{fontenc}
\usepackage{latexsym}
\usepackage{enumitem}
\usepackage{url}
\usepackage{amsmath}
\usepackage[nameinlink]{cleveref}
\crefformat{section}{\S#2#1#3} 
\crefformat{subsection}{\S#2#1#3}
\crefformat{subsubsection}{\S#2#1#3}
\crefformat{paragraph}{\S#2#1#3}
\usepackage{subcaption}
\usepackage{xspace}
\usepackage{graphicx}
\usepackage[most]{tcolorbox}
\usepackage{etoolbox}
\usepackage{algpseudocode}
\usepackage{multirow}
\usepackage{multicol}
\usepackage{tabularx}
\usepackage{mdframed}
\usepackage{enumitem}
\usepackage{booktabs}

\usepackage[utf8]{inputenc}
\usepackage{microtype}

\newcommand\GB[1]{\colorbox{green!30} {#1}}

\newcommand{\bertbu}{\text{BERT\textsubscript{base-uncased}}}
\newcommand{\bertbc}{\text{BERT\textsubscript{base-cased}}}
\newcommand{\srl}{\text{SRL}}
\newcommand{\ner}{\text{NER}}
\newcommand{\coref}{\text{CoREF}}
\newcommand{\pos}{\text{PoS}}
\newcommand{\squad}{SQuAD}
\newcommand{\record}{\text{ReCoRD}}
\newcommand{\multirc}{\text{MultiRC}}
\newcommand{\hotpot}{\text{HotpotQA}}
\newcommand{\fscore}{\text{F1}}
\newcommand{\microfscore}{\text{Micro-F1}}
\newcommand{\finetuning}{\text{fine-tuning}}
\newcommand{\finetuned}{\text{fine-tuned}}
\newcommand{\finetunedencoder}{\text{fine-tuned-encoder}}
\newcommand{\frozenencoder}{\text{frozen encoder}}
\newcommand{\quoref}{\text{Quoref}}
\newcommand{\modelagnosticfilter}{\text{Model-Agnostic Filter}}
\newcommand{\modeldependentfilter}{\text{Model-Dependent Filter}}
\newcommand{\unkentity}{\texttt{UNK\_ETYPE}}

\title{Can Edge Probing Tests Reveal Linguistic Knowledge in QA Models?}

\author[1*, 3]{Sagnik Ray Choudhury\Thanks{\enspace * Work done at the University of Copenhagen.}}
\author[2]{Nikita Bhutani}
\author[3]{Isabelle Augenstein}
\affil[1]{University of Michigan}
\affil[2]{Megagon Labs}
\affil[3]{University of Copenhagen}
\affil[ ]{\tt{sagnikrayc@gmail.com}, \tt{nikita@megagon.ai}, \tt{augenstein@di.ku.dk}}
\date{}
\makeatletter
\def\thanks#1{\protected@xdef\@thanks{\@thanks
        \protect\footnotetext{#1}}}
\makeatother

\begin{document}
\maketitle
\begin{abstract}
There have been many efforts to try to understand what grammatical knowledge (e.g., ability to understand the part of speech of a token) is encoded in large pre-trained language models (LM). 
This is done through `Edge Probing' (EP) tests: supervised classification tasks to predict the grammatical properties of a span (whether it has a particular part of speech) using \textit{only} the token representations coming from the LM encoder.
However, most NLP applications fine-tune these LM encoders for specific tasks. Here, we ask: if an LM is fine-tuned, does the encoding of linguistic information in it change, as measured by EP tests?
Specifically, we focus on the task of Question Answering (QA) and conduct experiments on multiple datasets. 
We find that EP test results do not change significantly when the fine-tuned model performs well or in adversarial situations where the model is forced to learn wrong correlations.
 From a similar finding, some recent papers conclude that fine-tuning does not change linguistic knowledge in encoders but they do not provide an explanation.
 We find that EP models themselves are susceptible to exploiting spurious correlations in the EP datasets. 
 When this dataset bias is corrected, we do see an improvement in the EP test results as expected.

\end{abstract}

\section{Introduction}

\label{sec:intro}

The encoding of linguistic information in large pre-trained language models (LMs) such as BERT \cite{DBLP:conf/naacl/DevlinCLT19} has become an active research topic in recent times. This encoding is usually measured by edge probing (EP) tasks \cite{DBLP:conf/naacl/Liu0BPS19, tenney-etal-2019-bert}. Consider the sentence ``the Met is closing soon''. The token `met' is a noun (a museum and not a form of the verb `meet'). The context words (`the', `is') are the only signals for determining its part of speech. If a `simple' (one or two layer MLP \cite{DBLP:conf/emnlp/HewittL19}) classifier predicts `met' as a noun \textit{only} using the representation of the token `met' (coming from a contextual encoder such as BERT \cite{DBLP:conf/naacl/DevlinCLT19} or ELMo \cite{peters-etal-2018-deep}) and not the whole sentence, then these signals must have been encoded in the token representation itself. This is the grammatical knowledge the test is `probing' for. If encoder $E_1$ performs better than encoder $E_2$ on an EP test, say, part-of-speech tagging, we say that $E_1$ has a better knowledge of part-of-speech than $E_2$. 

For many NLP tasks, pre-trained LMs (most commonly, BERT) have emerged as standard encoders \cite{JMLR:v21:20-074}. These encoders are \finetuned\ after adding task-specific layers on top. While probing tests on pre-trained encoders are quite popular, fine-tuned encoders are relatively under-explored (with notable exceptions of \citet{merchant-etal-2020-happens} and \citet{10.1145/3357384.3358028}. We aim to bridge this gap by probing \finetuned\ models using question answering (QA) as a target task. QA is a complex NLU problem requiring the model to implicitly perform many reasoning steps, and \finetuned\ models provide strong baselines for various QA datasets \cite{DBLP:conf/naacl/DevlinCLT19}. Our first research question is thus:

\textbf{RQ1}: \textit{Does \finetuning\ for QA tasks improve the encoding of linguistic skills in the encoders, when measured by existing EP tests?} 
Intuitively, DNN based QA models would require implicit knowledge of semantic roles (who did what to whom, when, and where), an understanding of the part of speech and entity boundaries (most answers are entities in the context), and anaphora resolution (entities in the context would refer to each other). 
Indeed, prior works show how injecting knowledge about semantic roles \cite{shen-lapata-2007-using} and coreference resolution \cite{vicedo-ferrandez-2000-importance} in classical QA pipelines improves their performance.
Therefore, a \finetuned\ QA model should \textit{implicitly} acquire these linguistic skills. 
The QA layers in the \finetuned\ models have much fewer parameters than the encoders (\cref{sec:epqa}), therefore, the encoders themselves are more likely to encode that grammatical knowledge. Consequently, when these \finetuned\ encoders are used for the \srl\ (semantic role labeling), \coref\ (coreference), \pos\ (part-of-speech tagging), and \ner\ (named entity recognition) EP tests, we would expect to see improvements over pre-trained LMs. \textbf{But we do not observe any such change (\cref{sec:rq1})}.
 
Fine-tuning is generally performed on much less data than pre-training and the encoder weights might not change significantly. Could that cause the EP test results to remain the same? If the encoder weights are kept fixed during \finetuning, the performance in the target (QA) task drops $50-70\%$ on all datasets. However, this frozen encoder has the same performance on the EP test as the original one. In other words, two encoders with a high difference in the target task performances have no discernible difference in the EP task performances. A possible explanation is that the QA models have no need to use the grammatical knowledge we are testing for. This motivates the second research question:

\textbf{RQ2}: \textit{Does \finetuning\ for QA tasks impart the linguistic skills necessary to perform QA in the encoders?} To answer this, we create a QA dataset that requires a particular `skill' \cite{DBLP:journals/corr/abs-2107-12708,choudhury2022machine}: the knowledge of coreference resolution (\cref{sec:rq2}). \quoref\ \cite{DBLP:conf/emnlp/DasigiLMSG19} is such a dataset, but one might not require the knowledge of coreference to answer \textit{all} questions in \quoref. Many instances in standard NLU tasks can be solved by heuristics, i.e., without proper \textit{reasoning} (see \citet{DBLP:conf/naacl/GururanganSLSBS18} for NLI or \citet{DBLP:conf/acl/MinWSGHZ19} for multi-hop QA). We design algorithms to filter out questions that can be answered heuristically (\cref{sec:rq2}), and consequently, any model \textit{probably} needs to use the knowledge of coreference to answer the rest. However, two encoders with a significant performance difference on this de-biased dataset have no difference in the \coref\ EP test. This motivates the third research question: 

\textbf{RQ3}: \textit{Why do the EP test results not reflect that encoders have learned the linguistic skills needed to perform QA?}: Our analysis (\cref{sec:rq3}) of the EP test datasets suggests that the EP models themselves might rely on dataset biases (as opposed to learning the task with input representations). When this bias is corrected, \finetuned\ encoders behave as expected, i.e., show significant performance improvements over the base encoders. Previously, \citet{10.1145/3357384.3358028} and \citet{merchant-etal-2020-happens} observed that the EP test results do not differ in the base vs \finetuned\ encoders\footnote{\citet{merchant-etal-2020-happens} uses one QA dataset (\squad~\cite{DBLP:conf/emnlp/RajpurkarZLL16}) and all our EP tests; \citet{10.1145/3357384.3358028} uses two datasets (\squad\ and \hotpot~\cite{DBLP:conf/emnlp/Yang0ZBCSM18}) and two of our EP tests; which makes this study more rigorous in the QA domain.} ($RQ1$) and concluded that the encoding of grammatical knowledge in the encoders does not change during \finetuning. However, unlike ours, their studies were not done on the problems that explicitly call for such grammatical knowledge. Moreover, current criticisms of EP tests on non \finetuned\ encoders focus on the task design itself \cite{DBLP:conf/emnlp/HewittL19, DBLP:conf/emnlp/VoitaT20} (see \cref{sec:rq3}), whereas this work calls for bias correction in the standardized EP test datasets.

\section{Related Work}

Prior work has focused on understanding various aspects of pre-trained LMs including attention patterns \cite{DBLP:conf/blackboxnlp/ClarkKLM19} and linguistic knowledge \cite{DBLP:conf/naacl/Liu0BPS19}. When these LMs are used as encoders in models, they turn out to be strong baselines for many tasks \cite{JMLR:v21:20-074}. However, less is known about how the \finetuning\ process changes the encoder's attention patterns \cite{kovaleva-etal-2019-revealing} or their encoding of linguistic knowledge. While many papers \cite[][inter alia]{DBLP:conf/emnlp/JiaL17, DBLP:conf/emnlp/KaushikL18, DBLP:conf/emnlp/SenS20, DBLP:conf/aaai/SugawaraSIA20} argue that DNN models often use heuristics to answer questions, \citet{choudhury2022machine} shows that at least some of the models use human-interpretable reasoning steps. Therefore, it is important to study how edge probing tests capture the task-specific reasoning abilities introduced in the \finetuning\ process.

The paradigm of the classifier based probing tasks (of which our EP tasks are a subset) is quite mature \cite{DBLP:conf/repeval/EttingerER16}, and has seen increasing popularity with the release of benchmark EP datasets (the ones we use here) \cite{tenney-etal-2019-bert}. Typically, internal layers of large language or machine translation models are used as features for auxiliary prediction tasks for syntactic properties: part-of-speech \cite{DBLP:conf/emnlp/ShiPK16, DBLP:conf/acl/BlevinsLZ18, tenney-etal-2019-bert}, tense \cite{DBLP:conf/emnlp/ShiPK16, tenney-etal-2019-bert}, or subject-verb agreement \cite{DBLP:conf/emnlp/TranBM18, DBLP:journals/tacl/LinzenDG16}. See \citet{DBLP:journals/tacl/BelinkovG19} for an extensive survey.

However, not many papers study the benchmark EP tests for \finetuned\ representations. Most similar to our work is the layer-wise analysis of BERT weights for QA \cite{10.1145/3357384.3358028} and the results of \finetuning\ BERT on EP tasks \cite{merchant-etal-2020-happens} -- \citet{10.1145/3357384.3358028} use three QA datasets (\squad, \hotpot, and bAbi \cite{DBLP:journals/corr/WestonBCM15}) to show that: 1) for the EP task of \coref, test results remain unchanged, even when representations are taken from different layers; 2) different layers of a \finetuned\ BERT can be attributed to different tasks in the QA process such as supporting fact extraction or entity selection. \citet{merchant-etal-2020-happens} studies MNLI, dependency parsing, and QA (\squad) to arrive at a similar finding, although their main results use a scalar mix of the weights from all layers of a fine-tuned BERT (whereas our work uses the top layer). RQ1 in our work can be considered complementary to their work, but RQ2 and RQ3 have not been studied before. 

EP tests are indirect, i.e., a classifier (probe) is used to measure the linguistic information in the representation. Do the test results reflect the quality of the representations or the classifier's ability to learn the task \cite{DBLP:conf/emnlp/HewittL19, DBLP:conf/emnlp/VoitaT20}? We discuss this in \cref{sec:rq3}. See \citet{DBLP:journals/corr/abs-2102-12452} for more background on probing classifiers.

\section{Edge Probing \& QA: The Setup}
\label{sec:epqa}

\begin{figure}[t]
  \centering
        \includegraphics[scale=0.48]{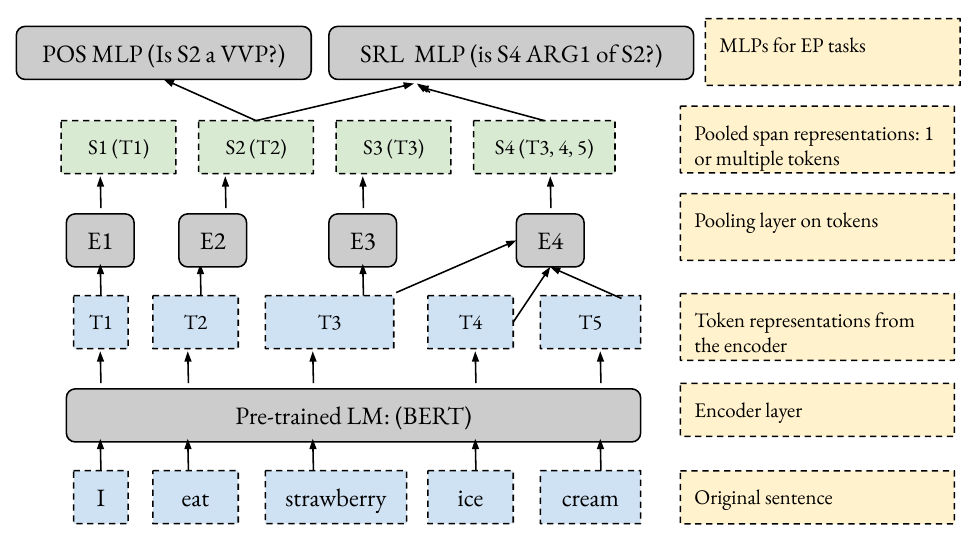}
    \caption{The architecture for edge probing tasks. For all tasks (\pos, \srl, \coref, \ner) the same MLP is used.} 
    \label{fig:ep-arch}
\end{figure}

\textbf{Edge Probing}: Following prior work \cite{merchant-etal-2020-happens, 10.1145/3357384.3358028}, we use the model architecture and four of the edge probing tasks proposed by \citet{tenney-etal-2019-bert}. 
Given a sentence $S = [T_{1},...T_{n}]$ of n tokens where $T_{i} \in \mathcal{R}^d$, for \pos\ and \ner\ tasks, the goal is to predict the part of speech or entity tag for a set of spans $T_{i}..T_{j}, 1 \le i, j \le n$ in that sentence (with \textit{only} the span and not the sentence as the input). 
Using the same setting for \srl\ and \coref\ tasks, the input is a pair of spans and the output is a class label: for \srl, it is a semantic role, for \coref\ it is a binary label indicating whether one span is an antecedent of the other or not. 
A self-attention pooling operator is used to generate a fixed representation for spans of different lengths \cite{lee-etal-2017-end}. 
For \srl\ and \coref, these representations for the two spans are concatenated. 
A single-layer linear probe is used for the actual classification task (\autoref{fig:ep-arch}).

For EP tests, the span representations need to be generated from the token representations. 
Token representations can be generated from each layer \cite{tenney-etal-2019-bert} or the top layer \cite{DBLP:conf/iclr/TenneyXCWPMKDBD19} of the encoder. 
In each case, the layer $i$ representation can be calculated as: 1. \textit{Just} the output of layer $i$, 2. A concatenation of the first layer output and layer $i$ output (`cat'), and 3. A scalar mixing of the output of $0-i$ layers where the mixing weights are trainable parameters (`mix'). In all experiments, we use the `cat' setting for the topmost layer in the encoder because both \citet{merchant-etal-2020-happens} and we find no significant difference for the other settings (the top layers generally perform better).

\noindent
\textbf{Question Answering}: In typical QA setups, models are given a context and a question as input. 
We use two span-based datasets (\squad, \citet{DBLP:conf/emnlp/RajpurkarZLL16} and \hotpot, \citet{DBLP:conf/emnlp/Yang0ZBCSM18}) where the task is to extract a span of the text from the context as the answer. We also use two MCQ datasets (\record, \citet{zhang2018record} and \multirc, \citet{khashabi-etal-2018-looking}) where the model is trained to select an answer from a set of choices. The architectures follow \citet{DBLP:conf/naacl/DevlinCLT19} and are similar for all datasets: one or two QA-specific layers on top of an encoder (see the appendix for more details).

\section{RQ1: EP Tests \& Fine-Tuned Encoders}
\label{sec:rq1}

We first run the EP tests with a standard encoder. Next, the QA models (the same encoder + QA layers) are trained (\finetuned).
Then, in the \finetuned\ models the QA layers are replaced with the same MLP layers used in the EP tests, and the tests are repeated. 
This gives us a measure of how much the encoding of linguistic knowledge in the encoder might have \textit{improved} due to the \finetuning\ process. 
We also randomize the QA training data (for \squad\ and \hotpot) by using random noun phrases as answers. Thus the model is forced to learn wrong correlations, which can cause it to `forget' the skills to some extent, and in turn, the encoder should perform worse on the EP tests. 

\subsection{Experimental Setup}

We use a \bertbu\ model as the base encoder for all tasks. For each QA task, we run five experiments. The span-based QA models are evaluated using the F1 score and the exact match (EM) metric. The exact match measures the percentage of answers that exactly match the actual answers. The F1 score measures the token overlap between the predicted and the actual answer. The MCQ questions are classification tasks and evaluated using the usual accuracy and \microfscore\ metrics for classification.

For EP tests we use the highest performing model in each QA dataset. The tests use the benchmark OntoNotes 5.0 corpus \cite{weischedel2013ontonotes}, as in \citet{tenney-etal-2019-bert} and \citet{merchant-etal-2020-happens}. We use the same hyper-parameters as the original paper on EP tests \cite{DBLP:conf/iclr/TenneyXCWPMKDBD19} except for the batch size ($32$ theirs vs $16$ ours).\footnote{\citet{merchant-etal-2020-happens} and \citet{10.1145/3357384.3358028} reportedly used the same HPs.} The QA models were trained for $10$ epochs with a batch size of $16$ using the Adam optimizer \cite{DBLP:journals/corr/KingmaB14}. The EP models were trained for three epochs, using the same batch size and optimizer. The learning rates were kept at \texttt{1e-04} for the EP tasks and \texttt{1e-05} for the \finetuning\ tasks. Further details about hyper-parameter searching and the exact configurations are provided in the appendix. Following the training regime of \cite{pruksachatkun2020jiant}, the models were evaluated on a subset of the validation data every $500$ mini-batches with early stopping on $100$ evaluations.

\subsection{Results}
We report the \microfscore\ scores for the EP tests on \finetuned\ models in \autoref{tab:ep-result} (the average over $5$ runs for each experiment, and the standard deviation varies between $0.1-1.5\%$). See the appendix for detailed results.

For \squad\, the test data is not publicly available, therefore, we report the results on dev data. We changed \hotpot\ instances to \squad\ style ones by providing the relevant sentences as the context, which is given for the train and dev data, but not for the test data. Therefore, we only report the results on the dev data.  

For \squad\ and \multirc, the results are somewhat lower than the best results reported in the literature with similar architectures ($88.5$ for \citet{DBLP:conf/naacl/DevlinCLT19} and $70.4$ for \citet{superglue-neurips}).\footnote{We use a max length of 128 in the encoder, whereas a max length of 512 produces comparable results. However, the target test results and EP test results are not correlated, therefore, we do not investigate this further.} For \record, the results are slightly better \cite{record-leaderboard}. For \hotpot, no fair evaluation is possible due to the data modifications.

Our EP test results for \srl\ and \coref\ do differ from the previous work \cite{tenney-etal-2019-bert}, but they are comparable with \citet{DBLP:conf/naacl/Liu0BPS19}, which uses the same dataset. However, we are more concerned with the fact that the EP test results do not change significantly when a \finetuned\ vs the original encoder is used. In the randomization experiments, we see that the QA \fscore\ score drops as expected, and the EP test results do change, but not as significantly as the QA \fscore\ scores. This also indicates that improving the performance of the QA model itself might not change the EP test results significantly.

In summary, our experiments suggest that \textbf{\finetuning\ indeed does not significantly change the encoding of linguistic knowledge in the underlying encoder, \textit{when measured by the EP tests}}, which is consistent with the findings of previous work \cite{merchant-etal-2020-happens, 10.1145/3357384.3358028}, but provides complementary evidence. 

\begin{table}[!t]
\footnotesize
\begin{tabular}{lrrrr}
\toprule
                             & \multicolumn{1}{l}{\srl} & \multicolumn{1}{l}{\coref} & \multicolumn{1}{l}{\pos} & \multicolumn{1}{l}{\ner} \\ \midrule
BERT-base                          & 81.1                     & 81.2                       & 96.1                     & 93.0                     \\
\midrule
\multicolumn{5}{l}{Fine-tuning on original data} \\ \midrule
\squad\ (81.9)  & 79.9                     & 81.2                       & 95.3                     & 92.4                    \\ 
\record\ (57.0)     & 79.7                     & 80.9                       & 95.8                    & 93.4                     \\
\multirc\ (63.7)    & 80.7                     & 82.3                       & 95.8                     & 93.5                     \\
\hotpot\ (77.0)             & 77.7                     & 80.2                       & 94.3                     & 90.9                     \\
\midrule
\multicolumn{5}{l}{Fine-tuning on randomized data} \\ \midrule
\squad\ (7.4) & 74.8                     & 78.9                       & 91.7                     & 86.8                     \\
\hotpot\ (12.5) & 74.0                     & 79.5                       & 92.0                     & 86.2                     \\
\bottomrule
\end{tabular}
\caption{\microfscore\ scores for different EP tasks: without fine-tuning, with fine-tuning on the original datasets, and with fine-tuning on randomized datasets. The \fscore\ scores for the QA datasets are given in parentheses.}
\label{tab:ep-result}
\end{table}

\section{RQ2: EP Test for Coreference}
 \label{sec:rq2}
 
 While we can expect the model to acquire some linguistic skills (the ability to do coreference resolution, identify the part of speech of a token) by learning to perform a QA task, there is no guarantee for this: a model can reason differently than we expect it to. For example, many \hotpot\ questions can be answered by identifying the necessary entity type and not the multi-hop reasoning process that we expect \cite{DBLP:conf/acl/MinWSGHZ19}.

Therefore, in $RQ2$, we want to see whether the EP test results change when we know the encoder has to acquire particular grammatical knowledge $K$ for the QA task. Consider two models $M_1$ ($E_1$ + $QA\_Layer_{1}$) and $M_2$ ($E_2$ + $QA\_Layer_{2}$) in our encoder + QA layer architecture. Assume we can identify a set of questions $Q$ that can \textit{only} be answered using $K$. If $M_1$ performs significantly better than $M_2$ in these questions, we can say that $E_1$ has encoded more information about $K$ than the $E_2$ because the QA layers are unlikely to encode that knowledge as they have much less parameters than the encoders. Therefore, in the EP test for $K$, we can expect $E_1$ to perform better than $E_2$. 

We define $M_1$ as a \textbf{\finetunedencoder}, where the full architecture (encoder ($E_1$) + QA layer) is trained; and $M_2$ as a \textbf{\frozenencoder}: the encoder ($E_2$) is frozen and \textit{only} the QA layer is trained. We choose $K$ to be the grammatical knowledge of coreference. It is difficult to understand whether the knowledge of semantic roles or part of speech would be needed to answer a question, but it possibly can be done for coreference. For example, in \autoref{fig:sample-squad}, it is relatively easy to see that to answer the question a human needs to resolve the reference `he' in the second sentence to `Leo Strauss'.

\begin{figure*}[!t]
    \centering
    \begin{subfigure}[b]{0.49\textwidth}
        \centering
        \begin{mdframed}[backgroundcolor=black!10,rightline=false,leftline=false]
        \footnotesize
        \textbf{Context}: \GB{Leo Strauss} ...was a German-American political philosopher ... 
         \GB{He} was born in Germany... Thoughts on Machiavelli is a book by \GB{Leo Strauss}... \\
        \textbf{Question}: Where was the author of Thoughts of Machiavelli born? \\
        \textbf{Answer}: Germany
        \end{mdframed}
        \caption{A sample question from \squad.}
         \label{fig:sample-squad}
    \end{subfigure}
    \hfill
    \begin{subfigure}[b]{0.49\textwidth}
        \centering
        \begin{mdframed}[backgroundcolor=black!10,rightline=false,leftline=false]
        \footnotesize
         \textbf{Context}:  \GB{Frankie} Bono, a mentally disturbed hitman from Cleveland,..Next, \GB{he} goes to purchase a revolver from Big Ralph....   \\
        \textbf{Question}: What is the first name of the person who purchases a revolver? \\
        \textbf{Answer}: Frankie
        \end{mdframed}
        \caption{A sample question from \quoref.}
        \label{fig:sample-quoref}
    \vspace{.4em}
    \end{subfigure}
    \caption{Sample questions from \squad\ and \quoref\ datasets. A reader relying on coreference resolution would take into account the \GB{green} tokens.}
\end{figure*}

\subsection{Finding Coreference Questions}
\label{sec:finding-coref-questions}

We employed four NLP practitioners to annotate 200 questions (100 each from \squad\ and \hotpot). Each annotator was given a sample of 100 questions and was asked to stop as soon as they found 50 positive (questions they thought required coreference) instances. The question in \autoref{fig:sample-squad} is sampled from that dataset. But the question can also be answered by a shortcut \cite{DBLP:journals/corr/abs-2004-07780}. We know a `where' question will only be answered by an entity of type location and there are two such entities in the context: Germany and United States. Germany is an argument to the trigger verb `born', hence, is the answer. 

The \quoref\ dataset \cite{DBLP:conf/emnlp/DasigiLMSG19} reportedly consists of questions that can only be answered by understanding the concept of coreference. The annotators \textit{design} the questions themselves, which is different from our post-hoc annotation process. \autoref{fig:sample-quoref} shows a sample question from the dataset. This question can not be answered without resolving the pronominal antecedent `he' to `Frankie'. But even \quoref\ \textit{can} have questions that can be answered with a shortcut, therefore we develop algorithms to filter them out (\cref{sec:model-agnostic-filter}, \cref{sec:model-dependent-filter}).

\subsubsection{\modelagnosticfilter}
\label{sec:model-agnostic-filter}

In \quoref, the answer is a span in the context. The \modelagnosticfilter\ algorithm works in two steps: a) \textbf{Sentence Selection}: Select the context sentence that is most similar to the question; and b) \textbf{Entity Type Matching}: The question expects an entity of a particular type, eg: `where' $\rightarrow$ location, `who' $\rightarrow$ person. From the sentence selected in the last step, select an entity of the same type.

For `Sentence Selection', we use two methods: 1) \textbf{Token-Overlap}: Select the sentence that has the highest token overlap with the question tokens, and 2) \textbf{Sentence Encoder}: An off-the-shelf sentence encoder from \citet{reimers-gurevych-2019-sentence} trained on MS Marco \cite{DBLP:conf/nips/NguyenRSGTMD16}, a large scale dataset for answer passage retrieval (see the appendix for details).

For the `Entity Type Matching' step, we design both supervised and unsupervised algorithms to determine the type of the answer entity from the question. For the unsupervised algorithm, we define a map (see the appendix) with the `wh' words (who, when, whom) as the keys and the entity type as values (who $\leftarrow$ \texttt{PER}). The first `wh' word in the question determines the output. For example, for the question ``where was Plato born, who wrote Republic?'' it produces \texttt{LOC}. If no such word is found, it outputs an \unkentity. 

This unsupervised approach will predict the wrong entity type for questions such as ``Who won the World Cup in 2002?'' (\texttt{PER} instead of \texttt{LOC}). Therefore, we train and evaluate supervised classification models on the training split of the \quoref\ dataset.\footnote{Further divided into $70$(train)-$20$(dev)-$10$(test) splits} The label for a question is the entity type of the answer,\footnote{One of the $18$ types in \citet{DBLP:conf/conll/PradhanMXNBUZZ13}} as detected by an off-the-shelf entity extractor from \href{https://spacy.io}{Spacy}. If the answer is not a named entity, or our entity extractor fails to determine its type, the label is \unkentity.\footnote{Indeed, a significant number of questions are labeled as such: $33\%$, $34\%$, and $35\%$ in the train, dev, and test split of the data respectively.} We experiment with two architectures: 1) a \finetuned\ \bertbc\ model; and 2) a popular word convolutional model for sentence classification \cite{DBLP:conf/emnlp/Kim14} using three parallel filters and 300 dimensional Google News Word2Vec representations \cite{DBLP:journals/corr/abs-1301-3781}. More details about the data, model architectures, and training is provided in the appendix.

\subsubsection{\modeldependentfilter} 
\label{sec:model-dependent-filter}

Following \citet{DBLP:conf/aaai/SugawaraSIA20} we replace all pronouns from the context in a question with random strings of the same length. If any one of $M_1$ or $M_2$ can still answer the question, it can \textit{arguably} be answered without the knowledge of coreference.

\subsubsection{Experiments \& Results}

The BERT and the WordConv supervised entity detectors have an average accuracy of $63.55 \pm 0.1 \%$, and $58.81 \pm 0.3 \%$ over 5 runs respectively.

\begin{table}[!t]
\centering
\footnotesize
\begin{tabular}{l l l l}
\toprule
Sentence                 & \multicolumn{2}{l}{Etype}                            & EM \\ \midrule
\multirow{3}{*}{Overlap} & \multirow{2}{*}{Supervised} & \finetuned\ (63) & 6.31        \\
                         &                             & WordConv (58)        & 6.27        \\ \cmidrule{2-4}
                         & \multicolumn{2}{l}{Unsupervised}             & 1.22        \\ \midrule
\multirow{3}{*}{Encoder} & \multirow{2}{*}{Supervised} & \finetuned\         & 5.99        \\
                         &                             & WordConv                & 5.48        \\ \cmidrule{2-4}
                         & \multicolumn{2}{l}{Unsupervised}                    & 0.97        \\ \bottomrule
\end{tabular}
\caption{Different strategies for the \modelagnosticfilter\ algorithm. EM stands for exact match, i.e., the percentage of cases where the filter produces the exact answer.}
\label{tab:model-agnostic-filter}
\end{table}

\begin{table}[!t]
\centering
\footnotesize
\begin{tabular}{l l l l l}
\toprule
        & \multicolumn{2}{l}{frozen} & \multicolumn{2}{l}{\finetuned} \\ \cmidrule{2-5}
          & \fscore           & EM           & \fscore            & EM            \\
\quoref\ dev     & 10.23        & 5.41        & 69.53          & 65.61         \\
- MAF      & 10.09        & 5.36        & 69.21          & 65.31         \\
- MDF     & 7.00        & 3.19        & 38.57          & 30.85         \\
- (MAF + MDF) & 6.76        & 2.97        & 38.38          & 30.69         \\ \midrule
\coref\ (\microfscore) &  \multicolumn{2}{l}{81.68 $\pm$ 1.68}       &  \multicolumn{2}{l}{83.11 $\pm$ 0.7}          \\ \bottomrule
\end{tabular}
\caption{Performance of \frozenencoder\ and \finetunedencoder\ models when filters are applied: separately and in combination. MAF and MDF stands for model agnostic and dependent filters. The last row reports the \microfscore\ for both encoders in \coref\ EP test.}
\label{tab:filter-results}
\end{table}

The `EM' column in \autoref{tab:model-agnostic-filter} shows the proportion of \quoref\ questions (the validation split) that can be answered by the \modelagnosticfilter\ algorithm. `Overlap' and `Encoder' are the two strategies for the `Sentence Selection' step, and `Supervised' and `Unsupervised' are the same for the `Entity Type Matching'. 

The final \modelagnosticfilter\ algorithm uses the token overlap approach to select a sentence from the context and uses the best \finetuned\ BERT model to find the entity type for the answer. With this, we can filter out $6.3\% (155/2418)$ questions from the dev set. While this number is not very high, a similar exercise on \squad\ determines that at least $21\%$ questions can be answered by this shortcut, which is consistent with prior findings \cite{DBLP:conf/emnlp/JiaL17}.

In the \modeldependentfilter\ algorithm, individually, the \finetuned\ model filters out $55\%$ dev instances, and the \frozenencoder\ model filters out $6\%$ of them, and in total, they filter out $56\%$ (there is a large overlap). 

\subsection{Target Task vs Encoder EP Test}
\label{sec:coref-fine-tuned-v-frozen}

\autoref{tab:filter-results} shows the results of the \finetuned\ and the frozen encoder on \quoref\ dev set before and after the filters are applied. As can be seen, the \finetuned\ encoder performs much better than the frozen one across all scenarios. The performance of these encoders on the \coref\ EP test is given in the last column, which, however, does not differ much. This proves that \textbf{while one encoder might encode a linguistic knowledge (coreference) better than the other, the EP tests might fail to capture that}.

\section{RQ3: An Analysis of EP Tests}
\label{sec:rq3}

\subsection{Analysis}
\label{sec:rq3-analysis}

We see that the EP test results are surprisingly stable, even when we expect the \finetuned\ encoder to learn or forget certain linguistic skills. EP tests are \textit{indirect} measures of a representation's quality and have been criticized as such. \citet{DBLP:conf/emnlp/VoitaT20} shows that for some EP tests, a large pre-trained LM (ElMo \cite{peters-etal-2018-deep}) has the same performance as a random encoder. They conclude that the test measures the classifiers' ability to learn the EP task, and not the knowledge encoded in the representations itself. They propose using an information-theoretic (minimum description length or MDL) probe. 

In a similar vein, \citet{DBLP:conf/emnlp/HewittL19} suggests designing a control task. A control task for an EP test is the same classification task, only the labels of the original task are changed so that it can not be recovered from the linguistic information. For example, two tokens with different part-of-speech tags will be mapped to the same arbitrary label. A classifier (probe) that performs well on both the control task and the original EP test must be learning the correlations in the data, and not using any information from the representations. Therefore, it can not measure the encoding of grammatical knowledge in the encoders.

If an EP test has only two labels, they will just be flipped in the control task. This makes the control task and the original EP test the same problem for the probe, and they must have the same performance. Therefore, for binary EP tests such as \coref\ (the one we are interested in), no control task can be designed, but we use simple linear probes following \citet{DBLP:conf/emnlp/HewittL19}'s recommendations. 

Had we used MDL probes instead, would our conclusions in $RQ1$ and $RQ2$ change? Prior work reports that the \finetuned\ encoders do not show much difference in the MDL probes themselves \cite{merchant-etal-2020-happens}. Moreover, even in the original MDL probe paper, the \coref\ test results are similar in a pre-trained vs random encoder \cite{DBLP:conf/emnlp/VoitaT20}. Therefore, replacing the current EP tests with MDL probes should not change the findings for the previous research questions.

Does this conclusively mean that the linguistic skill is not improved in the encoder, even when the task calls for it? Both \citet{merchant-etal-2020-happens} and \citet{10.1145/3357384.3358028} arrive at that conclusion, albeit without \finetuning\ on a skill-specific target dataset such as \quoref. We present a dataset bias explanation (\cref{sec:ep-heuristics}). Note the EP test dataset is used in both \citet{merchant-etal-2020-happens}, \citet{10.1145/3357384.3358028}, therefore the same explanation is valid for both these studies. 

\subsection{EP Test Heuristics}
\label{sec:ep-heuristics}

Following \citet{DBLP:conf/naacl/GururanganSLSBS18}, we design unsupervised algorithms that exploits spurious correlations in the dataset. 
\begin{itemize}[noitemsep, leftmargin=0pt, topsep=0pt]
    \item \textbf{Memorization}: If a test data point is in the training data, the classifier returns the training data label. Else, it returns a random label either a) uniformly sampled (`mem\_uniform') or b) sampled from the label probability distribution of the training data (`mem\_freq'). 
    \item \textbf{Same Precedent-Antecedent}: In the \coref\ dataset, whenever the precedent and antecedent are the same (``Obama is the president of the US. \textit{He} lives in Washington D.C. \textit{He} went to Harvard.''), return positive. 
\end{itemize}

\subsubsection{Results}
\begin{table}[!t]
\centering
\footnotesize
\begin{tabular}{l l l l l}
\toprule
               & \srl & \coref & \pos & \ner \\ \midrule
mem\_uniform       & 32.46               & 65.02                 & 88.62               & 71.59               \\
mem\_freq       & 44.45               & 78.06                 & 88.69               & 73.27               \\
same\_prec\_ante       & -               & 70.23\%                 & -              & -               \\

BERT-base      & 81.08 & 81.2 & 96.11 & 93.06    \\
\bottomrule
\end{tabular}
\caption{A performance comparison on EP test results: \microfscore\ scores for heuristics and \bertbu\ models (average over $5$ runs, STD. varies from $0.09-1\%$).}
\label{tab:baseline-comparison}
\end{table}

\begin{table}[!t]
\centering
\footnotesize
\begin{tabular}{l l l l}
\toprule
\multicolumn{2}{l}{}           & Best  & Worst \\ \midrule
\multirow{3}{*}{\srl}   & overall & - &  -7.05 \\ 
                       & easy    & - & -4.40 \\  
                       & hard    & - & -8.47 \\ \midrule
\multirow{3}{*}{\coref} & overall & +1.93 & -2.32\\ 
                       & easy    & \textbf{+4.51} & \textbf{-7.11} \\ 
                       & hard    & +0.47 & -0.17 \\ \midrule
\multirow{3}{*}{\pos}   & overall & - & -4.12 \\ \cmidrule{2-4} 
                       & easy    & - & -4.01 \\ 
                       & hard    & - & \textbf{-12.30} \\ \midrule
\multirow{3}{*}{\ner}   & overall & +0.35 & -6.85 \\ \cmidrule{2-4} 
                       & easy    & +0.25 & -4.82 \\ 
                       & hard    & +0.62 &  \textbf{-12.48}           \\ \midrule
\multirow{3}{*}{\coref-LS} & overall & +1.93 & -2.32\\ \cmidrule{2-4} 
                       & easy    & -1.22 & +1.1 \\ 
                       & hard    & \textbf{+13.27} & \textbf{-14.68} \\ 
                       \midrule
\end{tabular}
\caption{The accuracy changes across easy and hard instances for the best and worst \finetuned\ models. For \srl\ and \pos, BERT-base is the best, therefore, there is no positive change. \coref-LS refers to the case when the easy and hard points are created by splitting the data across labels. In some splits, the changes are significantly different than the overall change.}
\label{tab:ep-distr-mem}
\end{table}

\autoref{tab:baseline-comparison} shows the results for various heuristics and a \bertbu\ encoder. To achieve moderately high performance on most EP tests, no specific representation is needed, let alone from a pre-trained or a \finetuned\ one. The \textit{Same Precedent-Antecedent} heuristic has a \microfscore\ score of $70.23\%$ when used alone. This can be combined with mem\_freq/ mem\_uniform, but the combination provides no significant improvement. Overall, mem\_freq is a strong baseline for the EP tests.

\subsection{EP Tests: Hard \& Easy Instances}

The results in \cref{sec:ep-heuristics} can be viewed in another way. Many instances in the EP test data can be solved by memorization, i.e., are \textit{easy} instances. But there are definitely difficult ones that account for the performance difference in the heuristics and the \bertbu\ model. When the \finetuned\ encoders show marginal improvements or deterioration (over the base encoders) do the performance in these EP tests increase/decrease uniformly across the hard/easy instances, or in the harder instances the results change more drastically? \textbf{In the second case, it can be argued that the EP tests do actually capture the change in the encoders, but the change is artificially clamped, which is an unfortunate side-effect of the dataset}.

We first divide the test data for an EP test into easy and hard instances: the ones that can be solved by mem\_freq or not. Then we note the average accuracy of base \bertbu\ encoders on these splits. Finally, we take two QA models (from two separate datasets) for which the encoders had the average best/worst results in the said EP test. Do the results change from the base encoder similarly across these splits?

\subsubsection{Results}

\autoref{tab:ep-distr-mem} shows no discerning pattern in the `Best \finetuned' column, probably because the overall improvements are indeed not significant. However, when an encoder model performs poorly (the `Worst \finetuned' column), it performs disproportionately badly on the hard instances. This proves that while on the surface it might appear the results are similar \cite{merchant-etal-2020-happens, 10.1145/3357384.3358028}, they are indeed not. 

We are however more interested in \coref, because as discussed in $RQ2$, we have reasons to believe that some encoders have indeed learnt more about the skill of coreference than the others. However, contrary to the other results, it seems that the better/worse results come from the easy instances. 

\coref\ dataset has a significant label imbalance (compare row 1 and 2 in \autoref{tab:baseline-comparison}). A classifier predicting a negative label for all test instances can achieve an accuracy of $78.33\%$. If we split the data across the labels (\coref-LS), we see that the results change drastically, with a clear indication that a better encoder gets better by classifying the hard (positive) instances better, and a worse encoder fails harder on the same instances. Unsurprisingly, the better encoders come from the encoders trained on the \quoref\ dataset.

In summary, we show \textit{why} an EP model can fit well to the EP data without using a \textit{good} representation. This indicates that while \finetuning\ may improve the encoding of grammatical knowledge in encoders, the current EP tests (even the MDL probes) might not be able to capture it. \textbf{There are issues with the datasets rather than the task design itself}. This is a new explanation for the apparent consistency of EP test results in \finetuned\ models, whereas previous work has mostly focused on classifier knowledge (see \cref{sec:rq3-analysis}). 

\section{Conclusion and Future Work}
Edge probing tests are the predominant method to probe for linguistic information in large language models. We use them to evaluate how the process of \finetuning\ an LM for QA might change the grammatical knowledge in an encoder, and observe no significant differences between pre-trained and fine-tuned LMs. More importantly, we find this phenomenon in carefully designed target tasks where the models must use the said grammatical knowledge. 
From similar EP test results, previous works have concluded that \finetuning\ does not change the encoding of grammatical knowledge. 
However, our analysis provides a `dataset bias' explanation for the consistency of the results and provides some clues as to why \textit{any} representation tends to achieve very similar results for EP tests. This is different from the previous task-design criticisms of the EP tests. 

Do \finetuned\ NLU models score highly on benchmarks for the right reasons, i.e., follow the human reasoning process? This work shows some evidence in favor of that. The encoding of grammatical knowledge in QA encoders is improved as expected when the models are trained on the right datasets, and the dataset biases in the EP tests are corrected. In light of this evidence, in the future, we plan to identify the exact reasoning steps in the QA models through post-hoc explainability methods and study whether they align with the human reasoning steps.

\section{Acknowledgement}

We thank Behzad Golshan and Wang-Chiew Tan for the helpful discussions and annotation work. This project has been partly funded by a Megagon Labs research grant.

\bibliography{ep-qa-coling}

\begin{thebibliography}{49}
\expandafter\ifx\csname natexlab\endcsname\relax\def\natexlab#1{#1}\fi

\bibitem[{Belinkov(2022)}]{DBLP:journals/corr/abs-2102-12452}
Yonatan Belinkov. 2022.
\newblock \href {https://doi.org/10.1162/coli_a_00422} {{Probing Classifiers:
  Promises, Shortcomings, and Advances}}.
\newblock \emph{Computational Linguistics}, 48(1):207--219.

\bibitem[{Belinkov and Glass(2019)}]{DBLP:journals/tacl/BelinkovG19}
Yonatan Belinkov and James~R. Glass. 2019.
\newblock \href {https://doi.org/10.1162/tacl\_a\_00254} {{Analysis Methods in
  Neural Language Processing: {A} Survey}}.
\newblock \emph{Trans. Assoc. Comput. Linguistics}, 7:49--72.

\bibitem[{Blevins et~al.(2018)Blevins, Levy, and
  Zettlemoyer}]{DBLP:conf/acl/BlevinsLZ18}
Terra Blevins, Omer Levy, and Luke Zettlemoyer. 2018.
\newblock \href {https://doi.org/10.18653/v1/P18-2003} {{Deep RNNs Encode Soft
  Hierarchical Syntax}}.
\newblock In \emph{Proceedings of the 56th Annual Meeting of the Association
  for Computational Linguistics, {ACL} 2018, Melbourne, Australia, July 15-20,
  2018, Volume 2: Short Papers}, pages 14--19. Association for Computational
  Linguistics.

\bibitem[{Clark et~al.(2019)Clark, Khandelwal, Levy, and
  Manning}]{DBLP:conf/blackboxnlp/ClarkKLM19}
Kevin Clark, Urvashi Khandelwal, Omer Levy, and Christopher~D. Manning. 2019.
\newblock \href {https://doi.org/10.18653/v1/W19-4828} {{What Does {BERT} Look
  at? An Analysis of BERT's Attention}}.
\newblock In \emph{Proceedings of the 2019 {ACL} Workshop BlackboxNLP:
  Analyzing and Interpreting Neural Networks for NLP, BlackboxNLP@ACL 2019,
  Florence, Italy, August 1, 2019}, pages 276--286. Association for
  Computational Linguistics.

\bibitem[{Dasigi et~al.(2019)Dasigi, Liu, Marasovic, Smith, and
  Gardner}]{DBLP:conf/emnlp/DasigiLMSG19}
Pradeep Dasigi, Nelson~F. Liu, Ana Marasovic, Noah~A. Smith, and Matt Gardner.
  2019.
\newblock \href {https://doi.org/10.18653/v1/D19-1606} {{Quoref: A Reading
  Comprehension Dataset with Questions Requiring Coreferential Reasoning}}.
\newblock In \emph{Proceedings of the 2019 Conference on Empirical Methods in
  Natural Language Processing and the 9th International Joint Conference on
  Natural Language Processing, {EMNLP-IJCNLP} 2019, Hong Kong, China, November
  3-7, 2019}, pages 5924--5931. Association for Computational Linguistics.

\bibitem[{Devlin et~al.(2019)Devlin, Chang, Lee, and
  Toutanova}]{DBLP:conf/naacl/DevlinCLT19}
Jacob Devlin, Ming{-}Wei Chang, Kenton Lee, and Kristina Toutanova. 2019.
\newblock \href {https://doi.org/10.18653/v1/n19-1423} {{BERT: Pre-training of
  Deep Bidirectional Transformers for Language Understanding}}.
\newblock In \emph{Proceedings of the 2019 Conference of the North American
  Chapter of the Association for Computational Linguistics: Human Language
  Technologies, {NAACL-HLT} 2019, Minneapolis, MN, USA, June 2-7, 2019, Volume
  1 (Long and Short Papers)}, pages 4171--4186. Association for Computational
  Linguistics.

\bibitem[{Ettinger et~al.(2016)Ettinger, Elgohary, and
  Resnik}]{DBLP:conf/repeval/EttingerER16}
Allyson Ettinger, Ahmed Elgohary, and Philip Resnik. 2016.
\newblock \href {https://doi.org/10.18653/v1/W16-2524} {{Probing for semantic
  evidence of composition by means of simple classification tasks}}.
\newblock In \emph{Proceedings of the 1st Workshop on Evaluating Vector-Space
  Representations for NLP, RepEval@ACL 2016, Berlin, Germany, August 2016},
  pages 134--139. Association for Computational Linguistics.

\bibitem[{Geirhos et~al.(2020)Geirhos, Jacobsen, Michaelis, Zemel, Brendel,
  Bethge, and Wichmann}]{DBLP:journals/corr/abs-2004-07780}
Robert Geirhos, J{\"o}rn-Henrik Jacobsen, Claudio Michaelis, Richard Zemel,
  Wieland Brendel, Matthias Bethge, and Felix~A. Wichmann. 2020.
\newblock \href {https://doi.org/10.1038/s42256-020-00257-z} {Shortcut learning
  in deep neural networks}.
\newblock \emph{Nature Machine Intelligence}, 2(11):665--673.

\bibitem[{Gonz{\'{a}}lez and
  Rodr{\'{\i}}guez(2000)}]{vicedo-ferrandez-2000-importance}
Jos{\'{e}} Luis~Vicedo Gonz{\'{a}}lez and Antonio~Ferr{\'{a}}ndez
  Rodr{\'{\i}}guez. 2000.
\newblock \href {https://doi.org/10.3115/1075218.1075288} {{Importance of
  Pronominal Anaphora Resolution in Question Answering Systems}}.
\newblock In \emph{38th Annual Meeting of the Association for Computational
  Linguistics, Hong Kong, China, October 1-8, 2000}, pages 555--562. {ACL}.

\bibitem[{Gururangan et~al.(2018)Gururangan, Swayamdipta, Levy, Schwartz,
  Bowman, and Smith}]{DBLP:conf/naacl/GururanganSLSBS18}
Suchin Gururangan, Swabha Swayamdipta, Omer Levy, Roy Schwartz, Samuel~R.
  Bowman, and Noah~A. Smith. 2018.
\newblock \href {https://doi.org/10.18653/v1/n18-2017} {{Annotation Artifacts
  in Natural Language Inference Data}}.
\newblock In \emph{Proceedings of the 2018 Conference of the North American
  Chapter of the Association for Computational Linguistics: Human Language
  Technologies, NAACL-HLT, New Orleans, Louisiana, USA, June 1-6, 2018, Volume
  2 (Short Papers)}, pages 107--112. Association for Computational Linguistics.

\bibitem[{Hermann et~al.(2015)Hermann, Kocisk{\'{y}}, Grefenstette, Espeholt,
  Kay, Suleyman, and Blunsom}]{DBLP:conf/nips/HermannKGEKSB15}
Karl~Moritz Hermann, Tom{\'{a}}s Kocisk{\'{y}}, Edward Grefenstette, Lasse
  Espeholt, Will Kay, Mustafa Suleyman, and Phil Blunsom. 2015.
\newblock \href
  {https://proceedings.neurips.cc/paper/2015/hash/afdec7005cc9f14302cd0474fd0f3c96-Abstract.html}
  {{Teaching Machines to Read and Comprehend}}.
\newblock In \emph{Advances in Neural Information Processing Systems 28: Annual
  Conference on Neural Information Processing Systems 2015, December 7-12,
  2015, Montreal, Quebec, Canada}, pages 1693--1701.

\bibitem[{Hewitt and Liang(2019)}]{DBLP:conf/emnlp/HewittL19}
John Hewitt and Percy Liang. 2019.
\newblock \href {https://doi.org/10.18653/v1/D19-1275} {{Designing and
  Interpreting Probes with Control Tasks}}.
\newblock In \emph{Proceedings of the 2019 Conference on Empirical Methods in
  Natural Language Processing and the 9th International Joint Conference on
  Natural Language Processing, {EMNLP-IJCNLP} 2019, Hong Kong, China, November
  3-7, 2019}, pages 2733--2743. Association for Computational Linguistics.

\bibitem[{Jia and Liang(2017)}]{DBLP:conf/emnlp/JiaL17}
Robin Jia and Percy Liang. 2017.
\newblock \href {https://doi.org/10.18653/v1/d17-1215} {{Adversarial Examples
  for Evaluating Reading Comprehension Systems}}.
\newblock In \emph{Proceedings of the 2017 Conference on Empirical Methods in
  Natural Language Processing, {EMNLP} 2017, Copenhagen, Denmark, September
  9-11, 2017}, pages 2021--2031. Association for Computational Linguistics.

\bibitem[{Kaushik and Lipton(2018)}]{DBLP:conf/emnlp/KaushikL18}
Divyansh Kaushik and Zachary~C. Lipton. 2018.
\newblock \href {https://aclanthology.org/D18-1546/} {How much reading does
  reading comprehension require? {A} critical investigation of popular
  benchmarks}.
\newblock In \emph{Proceedings of the 2018 Conference on Empirical Methods in
  Natural Language Processing, Brussels, Belgium, October 31 - November 4,
  2018}, pages 5010--5015. Association for Computational Linguistics.

\bibitem[{Khashabi et~al.(2018)Khashabi, Chaturvedi, Roth, Upadhyay, and
  Roth}]{khashabi-etal-2018-looking}
Daniel Khashabi, Snigdha Chaturvedi, Michael Roth, Shyam Upadhyay, and Dan
  Roth. 2018.
\newblock \href {https://doi.org/10.18653/v1/N18-1023} {{Looking Beyond the
  Surface: A Challenge Set for Reading Comprehension over Multiple Sentences}}.
\newblock In \emph{Proceedings of the 2018 Conference of the North {A}merican
  Chapter of the Association for Computational Linguistics: Human Language
  Technologies, Volume 1 (Long Papers)}, pages 252--262, New Orleans,
  Louisiana. Association for Computational Linguistics.

\bibitem[{Kim(2014)}]{DBLP:conf/emnlp/Kim14}
Yoon Kim. 2014.
\newblock \href {https://doi.org/10.3115/v1/d14-1181} {{Convolutional Neural
  Networks for Sentence Classification}}.
\newblock In \emph{Proceedings of the 2014 Conference on Empirical Methods in
  Natural Language Processing, {EMNLP} 2014, October 25-29, 2014, Doha, Qatar,
  {A} meeting of SIGDAT, a Special Interest Group of the {ACL}}, pages
  1746--1751. {ACL}.

\bibitem[{Kingma and Ba(2015)}]{DBLP:journals/corr/KingmaB14}
Diederik~P. Kingma and Jimmy Ba. 2015.
\newblock \href {http://arxiv.org/abs/1412.6980} {{Adam: {A} Method for
  Stochastic Optimization}}.
\newblock In \emph{3rd International Conference on Learning Representations,
  {ICLR} 2015, San Diego, CA, USA, May 7-9, 2015, Conference Track
  Proceedings}.

\bibitem[{Kovaleva et~al.(2019)Kovaleva, Romanov, Rogers, and
  Rumshisky}]{kovaleva-etal-2019-revealing}
Olga Kovaleva, Alexey Romanov, Anna Rogers, and Anna Rumshisky. 2019.
\newblock \href {https://doi.org/10.18653/v1/D19-1445} {{Revealing the Dark
  Secrets of {BERT}}}.
\newblock In \emph{Proceedings of the 2019 Conference on Empirical Methods in
  Natural Language Processing and the 9th International Joint Conference on
  Natural Language Processing (EMNLP-IJCNLP)}, pages 4365--4374, Hong Kong,
  China. Association for Computational Linguistics.

\bibitem[{Lee et~al.(2017)Lee, He, Lewis, and Zettlemoyer}]{lee-etal-2017-end}
Kenton Lee, Luheng He, Mike Lewis, and Luke Zettlemoyer. 2017.
\newblock \href {https://doi.org/10.18653/v1/d17-1018} {{End-to-end Neural
  Coreference Resolution}}.
\newblock In \emph{Proceedings of the 2017 Conference on Empirical Methods in
  Natural Language Processing, {EMNLP} 2017, Copenhagen, Denmark, September
  9-11, 2017}, pages 188--197. Association for Computational Linguistics.

\bibitem[{Linzen et~al.(2016)Linzen, Dupoux, and
  Goldberg}]{DBLP:journals/tacl/LinzenDG16}
Tal Linzen, Emmanuel Dupoux, and Yoav Goldberg. 2016.
\newblock \href {https://doi.org/10.1162/tacl\_a\_00115} {Assessing the ability
  of lstms to learn syntax-sensitive dependencies}.
\newblock \emph{Trans. Assoc. Comput. Linguistics}, 4:521--535.

\bibitem[{Liu et~al.(2019)Liu, Gardner, Belinkov, Peters, and
  Smith}]{DBLP:conf/naacl/Liu0BPS19}
Nelson~F. Liu, Matt Gardner, Yonatan Belinkov, Matthew~E. Peters, and Noah~A.
  Smith. 2019.
\newblock \href {https://doi.org/10.18653/v1/n19-1112} {{Linguistic Knowledge
  and Transferability of Contextual Representations}}.
\newblock In \emph{Proceedings of the 2019 Conference of the North American
  Chapter of the Association for Computational Linguistics: Human Language
  Technologies, {NAACL-HLT} 2019, Minneapolis, MN, USA, June 2-7, 2019, Volume
  1 (Long and Short Papers)}, pages 1073--1094. Association for Computational
  Linguistics.

\bibitem[{Merchant et~al.(2020)Merchant, Rahimtoroghi, Pavlick, and
  Tenney}]{merchant-etal-2020-happens}
Amil Merchant, Elahe Rahimtoroghi, Ellie Pavlick, and Ian Tenney. 2020.
\newblock \href {https://doi.org/10.18653/v1/2020.blackboxnlp-1.4} {{What
  Happens To {BERT} Embeddings During Fine-tuning?}}
\newblock In \emph{Proceedings of the Third BlackboxNLP Workshop on Analyzing
  and Interpreting Neural Networks for NLP, BlackboxNLP@EMNLP 2020, Online,
  November 2020}, pages 33--44. Association for Computational Linguistics.

\bibitem[{Mikolov et~al.(2013)Mikolov, Chen, Corrado, and
  Dean}]{DBLP:journals/corr/abs-1301-3781}
Tom{\'{a}}s Mikolov, Kai Chen, Greg Corrado, and Jeffrey Dean. 2013.
\newblock \href {http://arxiv.org/abs/1301.3781} {{Efficient Estimation of Word
  Representations in Vector Space}}.
\newblock In \emph{1st International Conference on Learning Representations,
  {ICLR} 2013, Scottsdale, Arizona, USA, May 2-4, 2013, Workshop Track
  Proceedings}.

\bibitem[{Min et~al.(2019)Min, Wallace, Singh, Gardner, Hajishirzi, and
  Zettlemoyer}]{DBLP:conf/acl/MinWSGHZ19}
Sewon Min, Eric Wallace, Sameer Singh, Matt Gardner, Hannaneh Hajishirzi, and
  Luke Zettlemoyer. 2019.
\newblock \href {https://doi.org/10.18653/v1/p19-1416} {{Compositional
  Questions Do Not Necessitate Multi-hop Reasoning}}.
\newblock In \emph{Proceedings of the 57th Conference of the Association for
  Computational Linguistics, {ACL} 2019, Florence, Italy, July 28- August 2,
  2019, Volume 1: Long Papers}, pages 4249--4257. Association for Computational
  Linguistics.

\bibitem[{Nguyen et~al.(2016)Nguyen, Rosenberg, Song, Gao, Tiwary, Majumder,
  and Deng}]{DBLP:conf/nips/NguyenRSGTMD16}
Tri Nguyen, Mir Rosenberg, Xia Song, Jianfeng Gao, Saurabh Tiwary, Rangan
  Majumder, and Li~Deng. 2016.
\newblock \href {http://ceur-ws.org/Vol-1773/CoCoNIPS\_2016\_paper9.pdf} {{{MS}
  {MARCO:} {A} Human Generated MAchine Reading COmprehension Dataset}}.
\newblock In \emph{Proceedings of the Workshop on Cognitive Computation:
  Integrating neural and symbolic approaches 2016 co-located with the 30th
  Annual Conference on Neural Information Processing Systems {(NIPS} 2016),
  Barcelona, Spain, December 9, 2016}, volume 1773 of \emph{{CEUR} Workshop
  Proceedings}. CEUR-WS.org.

\bibitem[{Peters et~al.(2018)Peters, Neumann, Iyyer, Gardner, Clark, Lee, and
  Zettlemoyer}]{peters-etal-2018-deep}
Matthew Peters, Mark Neumann, Mohit Iyyer, Matt Gardner, Christopher Clark,
  Kenton Lee, and Luke Zettlemoyer. 2018.
\newblock \href {https://doi.org/10.18653/v1/N18-1202} {{Deep Contextualized
  Word Representations}}.
\newblock In \emph{Proceedings of the 2018 Conference of the North {A}merican
  Chapter of the Association for Computational Linguistics: Human Language
  Technologies, Volume 1 (Long Papers)}, pages 2227--2237, New Orleans,
  Louisiana. Association for Computational Linguistics.

\bibitem[{Pradhan et~al.(2013)Pradhan, Moschitti, Xue, Ng, Bj{\"{o}}rkelund,
  Uryupina, Zhang, and Zhong}]{DBLP:conf/conll/PradhanMXNBUZZ13}
Sameer Pradhan, Alessandro Moschitti, Nianwen Xue, Hwee~Tou Ng, Anders
  Bj{\"{o}}rkelund, Olga Uryupina, Yuchen Zhang, and Zhi Zhong. 2013.
\newblock \href {https://aclanthology.org/W13-3516/} {{Towards Robust
  Linguistic Analysis using OntoNotes}}.
\newblock In \emph{Proceedings of the Seventeenth Conference on Computational
  Natural Language Learning, CoNLL 2013, Sofia, Bulgaria, August 8-9, 2013},
  pages 143--152. {ACL}.

\bibitem[{Pruksachatkun et~al.(2020)Pruksachatkun, Yeres, Liu, Phang, Htut,
  Wang, Tenney, and Bowman}]{pruksachatkun2020jiant}
Yada Pruksachatkun, Phil Yeres, Haokun Liu, Jason Phang, Phu~Mon Htut, Alex
  Wang, Ian Tenney, and Samuel~R. Bowman. 2020.
\newblock \href {https://doi.org/10.18653/v1/2020.acl-demos.15} {{jiant: A
  Software Toolkit for Research on General-Purpose Text Understanding Models}}.
\newblock In \emph{Proceedings of the 58th Annual Meeting of the Association
  for Computational Linguistics: System Demonstrations}, pages 109--117,
  Online. Association for Computational Linguistics.

\bibitem[{Raffel et~al.(2020)Raffel, Shazeer, Roberts, Lee, Narang, Matena,
  Zhou, Li, and Liu}]{JMLR:v21:20-074}
Colin Raffel, Noam Shazeer, Adam Roberts, Katherine Lee, Sharan Narang, Michael
  Matena, Yanqi Zhou, Wei Li, and Peter~J. Liu. 2020.
\newblock \href {http://jmlr.org/papers/v21/20-074.html} {{Exploring the Limits
  of Transfer Learning with a Unified Text-to-Text Transformer}}.
\newblock \emph{Journal of Machine Learning Research}, 21(140):1--67.

\bibitem[{Rajpurkar et~al.(2016)Rajpurkar, Zhang, Lopyrev, and
  Liang}]{DBLP:conf/emnlp/RajpurkarZLL16}
Pranav Rajpurkar, Jian Zhang, Konstantin Lopyrev, and Percy Liang. 2016.
\newblock \href {https://doi.org/10.18653/v1/d16-1264} {{S{Q}u{AD}: 100, 000+
  Questions for Machine Comprehension of Text}}.
\newblock In \emph{Proceedings of the 2016 Conference on Empirical Methods in
  Natural Language Processing, {EMNLP} 2016, Austin, Texas, USA, November 1-4,
  2016}, pages 2383--2392. The Association for Computational Linguistics.

\bibitem[{Ray~Choudhury et~al.(2022)Ray~Choudhury, Rogers, and
  Augenstein}]{choudhury2022machine}
Sagnik Ray~Choudhury, Anna Rogers, and Isabelle Augenstein. 2022.
\newblock {Machine Reading, Fast and Slow: When Do Models ``Understand''
  Language?}
\newblock In \emph{Proceedings of the 29th {{International Conference}} on
  {{Computational Linguistics}}}. {International Committee on Computational
  Linguistics}.

\bibitem[{Reimers and Gurevych(2019)}]{reimers-gurevych-2019-sentence}
Nils Reimers and Iryna Gurevych. 2019.
\newblock \href {https://doi.org/10.18653/v1/D19-1410} {Sentence-{BERT}:
  Sentence embeddings using {S}iamese {BERT}-networks}.
\newblock In \emph{Proceedings of the 2019 Conference on Empirical Methods in
  Natural Language Processing and the 9th International Joint Conference on
  Natural Language Processing (EMNLP-IJCNLP)}, pages 3982--3992, Hong Kong,
  China. Association for Computational Linguistics.

\bibitem[{Rogers et~al.(2022)Rogers, Gardner, and
  Augenstein}]{DBLP:journals/corr/abs-2107-12708}
Anna Rogers, Matt Gardner, and Isabelle Augenstein. 2022.
\newblock \href {http://arxiv.org/abs/2107.12708} {{{QA} Dataset Explosion: {A}
  Taxonomy of {NLP} Resources for Question Answering and Reading
  Comprehension}}.
\newblock \emph{Computing Surveys (CSUR)}, to appear.

\bibitem[{Sen and Saffari(2020)}]{DBLP:conf/emnlp/SenS20}
Priyanka Sen and Amir Saffari. 2020.
\newblock \href {https://doi.org/10.18653/v1/2020.emnlp-main.190} {What do
  models learn from question answering datasets?}
\newblock In \emph{Proceedings of the 2020 Conference on Empirical Methods in
  Natural Language Processing, {EMNLP} 2020, Online, November 16-20, 2020},
  pages 2429--2438. Association for Computational Linguistics.

\bibitem[{Shen and Lapata(2007)}]{shen-lapata-2007-using}
Dan Shen and Mirella Lapata. 2007.
\newblock \href {https://aclanthology.org/D07-1002/} {{Using Semantic Roles to
  Improve Question Answering}}.
\newblock In \emph{EMNLP-CoNLL 2007, Proceedings of the 2007 Joint Conference
  on Empirical Methods in Natural Language Processing and Computational Natural
  Language Learning, June 28-30, 2007, Prague, Czech Republic}, pages 12--21.
  {ACL}.

\bibitem[{Shi et~al.(2016)Shi, Padhi, and Knight}]{DBLP:conf/emnlp/ShiPK16}
Xing Shi, Inkit Padhi, and Kevin Knight. 2016.
\newblock \href {https://doi.org/10.18653/v1/d16-1159} {{Does String-Based
  Neural {MT} Learn Source Syntax?}}
\newblock In \emph{Proceedings of the 2016 Conference on Empirical Methods in
  Natural Language Processing, {EMNLP} 2016, Austin, Texas, USA, November 1-4,
  2016}, pages 1526--1534. The Association for Computational Linguistics.

\bibitem[{Sugawara et~al.(2020)Sugawara, Stenetorp, Inui, and
  Aizawa}]{DBLP:conf/aaai/SugawaraSIA20}
Saku Sugawara, Pontus Stenetorp, Kentaro Inui, and Akiko Aizawa. 2020.
\newblock \href {https://ojs.aaai.org/index.php/AAAI/article/view/6422}
  {{Assessing the Benchmarking Capacity of Machine Reading Comprehension
  Datasets}}.
\newblock In \emph{The Thirty-Fourth {AAAI} Conference on Artificial
  Intelligence, {AAAI} 2020, The Thirty-Second Innovative Applications of
  Artificial Intelligence Conference, {IAAI} 2020, The Tenth {AAAI} Symposium
  on Educational Advances in Artificial Intelligence, {EAAI} 2020, New York,
  NY, USA, February 7-12, 2020}, pages 8918--8927. {AAAI} Press.

\bibitem[{Tenney et~al.(2019{\natexlab{a}})Tenney, Das, and
  Pavlick}]{tenney-etal-2019-bert}
Ian Tenney, Dipanjan Das, and Ellie Pavlick. 2019{\natexlab{a}}.
\newblock \href {https://doi.org/10.18653/v1/p19-1452} {{BERT Rediscovers the
  Classical NLP Pipeline}}.
\newblock In \emph{Proceedings of the 57th Conference of the Association for
  Computational Linguistics, {ACL} 2019, Florence, Italy, July 28- August 2,
  2019, Volume 1: Long Papers}, pages 4593--4601. Association for Computational
  Linguistics.

\bibitem[{Tenney et~al.(2019{\natexlab{b}})Tenney, Xia, Chen, Wang, Poliak,
  McCoy, Kim, Durme, Bowman, Das, and
  Pavlick}]{DBLP:conf/iclr/TenneyXCWPMKDBD19}
Ian Tenney, Patrick Xia, Berlin Chen, Alex Wang, Adam Poliak, R.~Thomas McCoy,
  Najoung Kim, Benjamin~Van Durme, Samuel~R. Bowman, Dipanjan Das, and Ellie
  Pavlick. 2019{\natexlab{b}}.
\newblock \href {https://openreview.net/forum?id=SJzSgnRcKX} {What do you learn
  from context? probing for sentence structure in contextualized word
  representations}.
\newblock In \emph{7th International Conference on Learning Representations,
  {ICLR} 2019, New Orleans, LA, USA, May 6-9, 2019}. OpenReview.net.

\bibitem[{Tran et~al.(2018)Tran, Bisazza, and Monz}]{DBLP:conf/emnlp/TranBM18}
Ke~M. Tran, Arianna Bisazza, and Christof Monz. 2018.
\newblock \href {https://aclanthology.org/D18-1503/} {{The importance of Being
  Recurrent for Modeling Hierarchical Structure}}.
\newblock In \emph{Proceedings of the 2018 Conference on Empirical Methods in
  Natural Language Processing, Brussels, Belgium, October 31 - November 4,
  2018}, pages 4731--4736. Association for Computational Linguistics.

\bibitem[{van Aken et~al.(2019)van Aken, Winter, L{\"{o}}ser, and
  Gers}]{10.1145/3357384.3358028}
Betty van Aken, Benjamin Winter, Alexander L{\"{o}}ser, and Felix~A. Gers.
  2019.
\newblock \href {https://doi.org/10.1145/3357384.3358028} {{How Does {BERT}
  Answer Questions?: {A} Layer-Wise Analysis of Transformer Representations}}.
\newblock In \emph{Proceedings of the 28th {ACM} International Conference on
  Information and Knowledge Management, {CIKM} 2019, Beijing, China, November
  3-7, 2019}, pages 1823--1832. {ACM}.

\bibitem[{Voita and Titov(2020)}]{DBLP:conf/emnlp/VoitaT20}
Elena Voita and Ivan Titov. 2020.
\newblock \href {https://doi.org/10.18653/v1/2020.emnlp-main.14}
  {{Information-Theoretic Probing with Minimum Description Length}}.
\newblock In \emph{Proceedings of the 2020 Conference on Empirical Methods in
  Natural Language Processing, {EMNLP} 2020, Online, November 16-20, 2020},
  pages 183--196. Association for Computational Linguistics.

\bibitem[{Wang et~al.(2019)Wang, Pruksachatkun, Nangia, Singh, Michael, Hill,
  Levy, and Bowman}]{superglue-neurips}
Alex Wang, Yada Pruksachatkun, Nikita Nangia, Amanpreet Singh, Julian Michael,
  Felix Hill, Omer Levy, and Samuel~R. Bowman. 2019.
\newblock \href
  {https://proceedings.neurips.cc/paper/2019/hash/4496bf24afe7fab6f046bf4923da8de6-Abstract.html}
  {{SuperGLUE: {A} Stickier Benchmark for General-Purpose Language
  Understanding Systems}}.
\newblock In \emph{Advances in Neural Information Processing Systems 32: Annual
  Conference on Neural Information Processing Systems 2019, NeurIPS 2019,
  December 8-14, 2019, Vancouver, BC, Canada}, pages 3261--3275.

\bibitem[{Weischedel et~al.(2013)Weischedel, Palmer, Marcus, Hovy, Pradhan,
  Ramshaw, Xue, Taylor, Kaufman, Franchini et~al.}]{weischedel2013ontonotes}
Ralph Weischedel, Martha Palmer, Mitchell Marcus, Eduard Hovy, Sameer Pradhan,
  Lance Ramshaw, Nianwen Xue, Ann Taylor, Jeff Kaufman, Michelle Franchini,
  et~al. 2013.
\newblock Ontonotes release 5.0 ldc2013t19.
\newblock \emph{Linguistic Data Consortium, Philadelphia, PA}, 23.

\bibitem[{Weston et~al.(2016)Weston, Bordes, Chopra, and
  Mikolov}]{DBLP:journals/corr/WestonBCM15}
Jason Weston, Antoine Bordes, Sumit Chopra, and Tomas Mikolov. 2016.
\newblock \href {http://arxiv.org/abs/1502.05698} {{Towards AI-Complete
  Question Answering: {A} Set of Prerequisite Toy Tasks}}.
\newblock In \emph{4th International Conference on Learning Representations,
  {ICLR} 2016, San Juan, Puerto Rico, May 2-4, 2016, Conference Track
  Proceedings}.

\bibitem[{Yang et~al.(2018)Yang, Qi, Zhang, Bengio, Cohen, Salakhutdinov, and
  Manning}]{DBLP:conf/emnlp/Yang0ZBCSM18}
Zhilin Yang, Peng Qi, Saizheng Zhang, Yoshua Bengio, William~W. Cohen, Ruslan
  Salakhutdinov, and Christopher~D. Manning. 2018.
\newblock \href {https://doi.org/10.18653/v1/d18-1259} {{HotpotQA: {A} Dataset
  for Diverse, Explainable Multi-hop Question Answering}}.
\newblock In \emph{Proceedings of the 2018 Conference on Empirical Methods in
  Natural Language Processing, Brussels, Belgium, October 31 - November 4,
  2018}, pages 2369--2380. Association for Computational Linguistics.

\bibitem[{Zeiler(2012)}]{DBLP:journals/corr/abs-1212-5701}
Matthew~D. Zeiler. 2012.
\newblock \href {http://arxiv.org/abs/1212.5701} {{{ADADELTA:} An Adaptive
  Learning Rate Method}}.
\newblock \emph{CoRR}, abs/1212.5701.

\bibitem[{Zhang(2020)}]{record-leaderboard}
Sheng Zhang. 2020.
\newblock \href {https://sheng-z.github.io/ReCoRD-explorer/} {{BERT}
  fine-tuning re-implementation for {ReCoRD}, {JHU}}.

\bibitem[{Zhang et~al.(2018)Zhang, Liu, Liu, Gao, Duh, and
  Durme}]{zhang2018record}
Sheng Zhang, Xiaodong Liu, Jingjing Liu, Jianfeng Gao, Kevin Duh, and
  Benjamin~Van Durme. 2018.
\newblock \href {http://arxiv.org/abs/1810.12885} {{ReCoRD: Bridging the Gap
  between Human and Machine Commonsense Reading Comprehension}}.
\newblock \emph{CoRR}, abs/1503.06733.

\end{thebibliography}
\bibliographystyle{acl_natbib}

\clearpage
\appendix

\section{Appendix A: RQ1}

\subsection{QA Datasets and Model Architectures}

\textbf{\squad}: We use the first (\texttt{V1.1}) version of the \squad\ dataset, which contains around 100K question, answer, context triples collected from the English Wikipedia. The answers are spans within the context. The questions and contexts are concatenated, and a linear layer on top of a contextual encoder is used to predict the probability of a context token $i$ being the start ($P_{i, s}$) or end ($P_{i, e}$) of an answer. The score ($S_{i,j}$) for a span with start token $i$ and end token $j$ is $P_{i, s} + P_{j, e}$. For all valid combinations of $i$ and $j$, the span with the highest score is chosen as the answer. A cross-entropy loss between the actual and predicted start/end positions is minimized. \\
\textbf{\hotpot}: The \hotpot\ dataset is a collection of 113K question-answer pairs, with two improvements over \squad: 1) Each context consists of multiple paragraphs (as opposed to a single one) and the model needs to reason over some of them to provide an answer and 2) The sentences required to answer a question are provided (as paragraph index, sentence index). The dataset features an implicit IR task (finding the relevant sentences), therefore we reduce it to a \squad\ style one by changing the context to the full paragraphs from where the supporting facts are chosen and removing the questions where the answer spans cannot be found in the context (3.8 \% in train and 3.9 \% in dev data).\\
\textbf{\record}: The \record\ dataset contains a set of 120,000 Close style questions from the CNN/Daily Mail dataset \cite{DBLP:conf/nips/HermannKGEKSB15}. A Cloze style query is a statement with an occluded entity that is factually supported by a passage. The dataset provides the named entities in the context, one of which is the answer. For each (question, context, $N$ named entity) triple in the data, $N$ (question, context, label) triples are created, with the missing entity in the question replaced by one of the provided entities, and the label is put as true or false depending on whether the entity is the answer or not. This NLI style formulation reduces the QA task to a classification problem, for which a two-layer MLP is used on top of the encoder layer: a linear layer with a tanh activation, followed by another linear layer.\\
\textbf{\multirc}: \multirc\ is a multiple-choice QA dataset where the model has to choose one or multiple of provided answers utilizing the text from the question, context, and the answer itself. We reduce it to a binary classification task, where the input is a concatenation of the question, the context, and the answer (for each of the possible answers). The same architecture as \record\ is used.

\subsection{HP Searching and Final Configurations for QA Datasets and EP Tests}

\textbf{Training Details}: We searched for the following hyper-parameters (HPs): number of epochs and learning rates. Finally, the QA models were trained for 10 epochs with a batch size of 16 using the Adam optimizer \cite{DBLP:journals/corr/KingmaB14}. The EP models were trained for the three epochs, using the same batch size and optimizer. The learning rates were kept at \texttt{1e-04} for the EP tasks and \texttt{1e-05} for the \finetuning\ tasks. Following the training regime of \cite{pruksachatkun2020jiant}, the model was evaluated on a subset of the validation data every 500 mini-batches with early stopping on 100 evaluations. The config files in the provided code show the detailed HPs. For the EP tests, the hyper-parameters are all same as the baseline \cite{DBLP:conf/iclr/TenneyXCWPMKDBD19} except for the batch size (32 theirs vs 16 ours). \citet{merchant-etal-2020-happens} and \citet{10.1145/3357384.3358028} reportedly used the same HPs. 

\subsection{Reproducibility Checklist}
\textbf{Description of computing infrastructure used}: Titan RTX GPU, CUDA version 11.2. \\
\textbf{The average runtime for each model or algorithm (e.g., training, inference, etc.), or estimated energy cost}: 5-6 Hours for EP tests, 3-4 hours for QA models. \\
\textbf{Number of parameters in each model}: For QA models: BERT base uncased parameters + 128*num\_classes (FC layer). For EP tests, they are the same. \\
\textbf{Corresponding validation performance for each reported test result}: For three QA datasets, we use the validation data itself. For the EP models, validation results are very similar to the test results reported. \\
\textbf{Explanation of evaluation metrics used, with links to code}: The evaluation metric for QA models is \fscore, as common in most QA datasets, including \squad. For EP tests, the evaluation metric is \microfscore, which is used in \citet{DBLP:conf/iclr/TenneyXCWPMKDBD19}, the paper which is our baseline. The implementations are from \citet{pruksachatkun2020jiant}, which hosts the original code used in \citet{DBLP:conf/iclr/TenneyXCWPMKDBD19}. We also use 

For all experiments with hyperparameter search:

\textbf{The exact number of training and evaluation runs}: For each QA model, 5 training/evaluation runs. For each EP test, 5 training/eval run. \\
\textbf{Bounds for each hyperparameter}: training batch size: 8-32, learning rate: for QA models, \texttt{1e-05} to \texttt{1e-03}, 3 steps. For EP tests, \texttt{1e-06} to \texttt{1e-04}, 5 steps. \\
\textbf{Hyperparameter configurations for best-performing models}: Provided as config files. \\
\textbf{The method of choosing hyperparameter values (e.g., uniform sampling, manual tuning, etc.) and the criterion used to select among them (e.g., accuracy)}: Uniform sampling, \fscore\ for QA models in dev data, \microfscore\ for EP tests in test data. \\

\subsection{Results}

Detailed results for the experiments in \cref{sec:rq1} are provided below in Tables \ref{tab:srl}, \ref{tab:coref}, \ref{tab:pos}, and \ref{tab:ner}.\\

\begin{table}[!t]
\centering
\footnotesize
\begin{tabularx}{0.5\textwidth}{|X|X|X|X|X|}
\hline
model     & acc              & micro\_f1        & macro\_f1        & weighted \_f1     \\ \hline
BERT-base & 81.08 $\pm$ 1.33 & 81.08 $\pm$ 1.33 & 31.93 $\pm$ 1.34 & 80.09 $\pm$ 1.54 \\ \hline
\multicolumn{5}{|c|}{\finetuned\ on original data}                                      \\ \hline
\squad\     & 79.97 $\pm$ 1.1  & 79.97 $\pm$ 1.1  & 30.57 $\pm$ 1.0  & 78.86 $\pm$ 1.3  \\ \hline
\record\    & 79.71 $\pm$ 1.46 & 79.71 $\pm$ 1.46 & 30.47 $\pm$ 1.61 & 78.61 $\pm$ 1.7  \\ \hline
\multirc\   & 80.69 $\pm$ 1.25 & 80.69 $\pm$ 1.25 & 31.76 $\pm$ 1.37 & 79.68 $\pm$ 1.47 \\ \hline
Hotpot    & 77.73 $\pm$ 1.2  & 77.73 $\pm$ 1.2  & 28.4 $\pm$ 1.23  & 76.45 $\pm$ 1.43 \\ \hline
\multicolumn{5}{|c|}{finetuned on randomized data}                                    \\ \hline
\squad     & 74.79 $\pm$ 0.15 & 74.79 $\pm$ 0.15 & 26.07 $\pm$ 0.56 & 73.36 $\pm$ 0.17 \\ \hline
Hotpot    & 74.03 $\pm$ 0.33 & 74.03 $\pm$ 0.33 & 25.79 $\pm$ 0.38 & 72.5 $\pm$ 0.35  \\ \hline
\end{tabularx}
\caption{Results for \srl\ EP test.}
\label{tab:srl}
\end{table}

\begin{table}[!t]
\centering
\footnotesize
\begin{tabularx}{0.5\textwidth}{|X|X|X|X|X|}
\hline
model     & acc            & micro\_f1      & macro\_f1      & weighted \_f1   \\ \hline
BERT-base & 81.18 $\pm$ 1.68 & 81.18 $\pm$ 1.68 & 59.33 $\pm$ 7.99 & 76.2 $\pm$ 3.89  \\ \hline
\multicolumn{5}{|c|}{\finetuned\ on original data}                              \\ \hline
\quoref\     &   83.11 $\pm$ 0.7 &   83.11 $\pm$ 0.7 &  67.97 $\pm$ 2.05 &  80.45 $\pm$ 1.09 \\ \hline
\squad\     & 81.19 $\pm$ 1.49 & 81.19 $\pm$ 1.49 & 60.27 $\pm$ 8.48 & 76.58 $\pm$ 4.01 \\ \hline
\record\   & 80.88 $\pm$ 1.76 & 80.88 $\pm$ 1.76 & 58.08 $\pm$ 9.79 & 75.57 $\pm$ 4.66 \\ \hline
\multirc\   & 82.37 $\pm$ 1.71 & 82.37 $\pm$ 1.71 & 65.12 $\pm$ 8.7  & 78.99 $\pm$ 4.19 \\ \hline
Hotpot    & 80.19 $\pm$ 1.84 & 80.19 $\pm$ 1.84 & 55.38 $\pm$ 8.86 & 74.21 $\pm$ 4.31 \\ \hline
\multicolumn{5}{|c|}{finetuned on randomized data}                            \\ \hline
\squad\     & 78.86 $\pm$ 0.38 & 78.86 $\pm$ 0.38 & 48.77 $\pm$ 3.29 & 71.01 $\pm$ 1.5  \\ \hline
Hotpot    & 79.45 $\pm$ 0.23 & 79.45 $\pm$ 0.23 & 52.1 $\pm$ 2.31  & 72.61 $\pm$ 1.03 \\ \hline
\end{tabularx}
\caption{Results for \coref\ EP test}
\label{tab:coref}
\end{table}

\begin{table}[!t]
\centering
\footnotesize
\begin{tabularx}{0.5\textwidth}{|X|X|X|X|X|}
\hline
model     & acc            & micro\_f1      & macro\_f1      & weighted \_f1   \\ \hline
BERT-base & 96.11 $\pm$ 0.15 & 96.11 $\pm$ 0.15 & 87.88 $\pm$ 0.8  & 96.06 $\pm$ 0.15 \\ \hline
\multicolumn{5}{|c|}{\finetuned\ on original data}                              \\ \hline
\squad\     & 95.27 $\pm$ 0.04 & 95.27 $\pm$ 0.04 & 87.08 $\pm$ 0.45 & 95.2 $\pm$ 0.05  \\ \hline
\record\    & 95.77 $\pm$ 0.08 & 95.77 $\pm$ 0.08 & 86.91 $\pm$ 0.39 & 95.71 $\pm$ 0.09 \\ \hline
\multirc\   & 95.77 $\pm$ 0.19 & 95.77 $\pm$ 0.19 & 86.51 $\pm$ 1.29 & 95.71 $\pm$ 0.19 \\ \hline
Hotpot    & 94.31 $\pm$ 0.09 & 94.31 $\pm$ 0.09 & 83.45 $\pm$ 0.7  & 94.21 $\pm$ 0.1  \\ \hline
\multicolumn{5}{|c|}{\finetuned\ on randomized data}                            \\ \hline
\squad\     & 91.74 $\pm$ 0.35 & 91.74 $\pm$ 0.35 & 78.52 $\pm$ 1.15 & 91.54 $\pm$ 0.37 \\ \hline
Hotpot    & 91.99 $\pm$ 0.13 & 91.99 $\pm$ 0.13 & 79.59 $\pm$ 0.65 & 91.8 $\pm$ 0.14  \\ \hline
\end{tabularx}
\caption{Results for \pos\ EP test}
\label{tab:pos}
\end{table}

\begin{table}[!t]
\centering
\footnotesize
\begin{tabularx}{0.5\textwidth}{|X|X|X|X|X|}
\hline
model                   & acc            & micro\_f1      & macro\_f1      & weighted \_f1   \\ \hline
BERT-base               & 93.0 $\pm$ 0.28  & 93.0 $\pm$ 0.28  & 78.12 $\pm$ 1.1  & 92.29 $\pm$ 0.37 \\ \hline
\multicolumn{5}{|c|}{\finetuning\ on original data}             \\ \hline
\squad\   & 92.44 $\pm$ 0.09 & 92.44 $\pm$ 0.09 & 78.19 $\pm$ 0.47 & 91.86 $\pm$ 0.12 \\ \hline
\record\                  & 93.35 $\pm$ 0.24 & 93.35 $\pm$ 0.24 & 79.88 $\pm$ 0.8  & 92.79 $\pm$ 0.31 \\ \hline
\multirc\ & 93.5 $\pm$ 0.4   & 93.5 $\pm$ 0.4   & 80.51 $\pm$ 1.54 & 93.0 $\pm$ 0.52  \\ \hline
Hotpot  & 90.9 $\pm$ 0.16  & 90.9 $\pm$ 0.16  & 73.81 $\pm$ 0.62 & 89.94 $\pm$ 0.21 \\ \hline
\multicolumn{5}{|c|}{\finetuning\ on randomized data}           \\ \hline
\squad\   & 86.77 $\pm$ 1.02 & 86.77 $\pm$ 1.02 & 64.05 $\pm$ 3.64 & 85.25 $\pm$ 1.26 \\ \hline
Hotpot  & 86.15 $\pm$ 0.32 & 86.15 $\pm$ 0.32 & 61.56 $\pm$ 1.42 & 84.48 $\pm$ 0.37 \\ \hline
\end{tabularx}
\caption{Results for \ner\ EP test}
\label{tab:ner}
\end{table}

\section{Appendix B: RQ2}

\subsection{Sentence Embedding Model for Answer Sentence Selection}

MS MARCO \cite{DBLP:conf/nips/NguyenRSGTMD16} is a large dataset of question/ answer pairs. The dataset was built by first sampling queries from Bing search log and then using Bing to retrieve relevant documents and automatically extract relevant passages from those documents. Annotators were asked to mark relevant spans from the documents as answers.

We us a sentence embedding model \cite{reimers-gurevych-2019-sentence} built on the MS MARCO dataset. The pre-trained model is available off-the-shelf,\footnote{https://www.sbert.net/docs/pretrained-models/msmarco-v5.html} therefore can be used directly to find the most `similar' sentence to the question. Among many sentence embedding models available for semantic search, we use this one because it is specifically trained on a question-passage retrieval dataset using a bi-encoder model. During training, the questions and the relevant/non-relevant passages are passed through a contextual encoder and their pooled representations are compared. The model is trained with the following objective: the (query, positive\_passage) pair is supposed to be close in the vector space, while (query, negative\_passage) should be distant.

\subsection{Unsupervised Entity Type Selection}

In unsupervised entity type selection method, we use a map to determine the entity type of the answer for a question. The map is given below:

\texttt{
            \{ \\
                `how far': [QUANTITY], \\
                `how long': [DATE], \\
                `how many': [CARDINAL], \\
                `how old': [QUANTITY], \\
                `what': [PRODUCT, WORK\_OF\_ART], \\
                `when': [DATE, TIME], \\
                `where': [FAC, LOC, ORG, GPE], \\
                `who': [PERSON], \\
                `whom': [PERSON], \\
                `whose': [PERSON, ORG, NORP] \\
            \}
}

The entity types are defined in \citet{DBLP:conf/conll/PradhanMXNBUZZ13}. If the question has the phrase `how far', the returned entity type is QUANTITY. The map is an \texttt{OrderedDict}, i.e., if the question is `how old is the person who wrote Harry Potter', the returned entity type is QUANTITY. When there are multiple possibilities (`where'), one is returned randomly.

\subsection{Supervised Entity Type Selection}

\subsubsection{Dataset}

The dataset is created using the training set of \quoref\, which is divided into train/dev/test split for entity type detector model training and evaluation. A sample data point is shown in \autoref{fig:sample-classification}.

\begin{figure}[!t]
\footnotesize
\begin{mdframed}[backgroundcolor=black!8,rightline=false,leftline=false]
\textbf{Text}: What is the full name of the person who is the television reporter that brings in a priest versed in Catholic exorcism rites? \\
\textbf{Label}: PER \\
\end{mdframed}
    \caption{A sample instance for answer entity type classifier.} 
    \label{fig:sample-classification}
\end{figure}

\begin{figure*}[htp]
  \centering
        \includegraphics[scale=0.5]{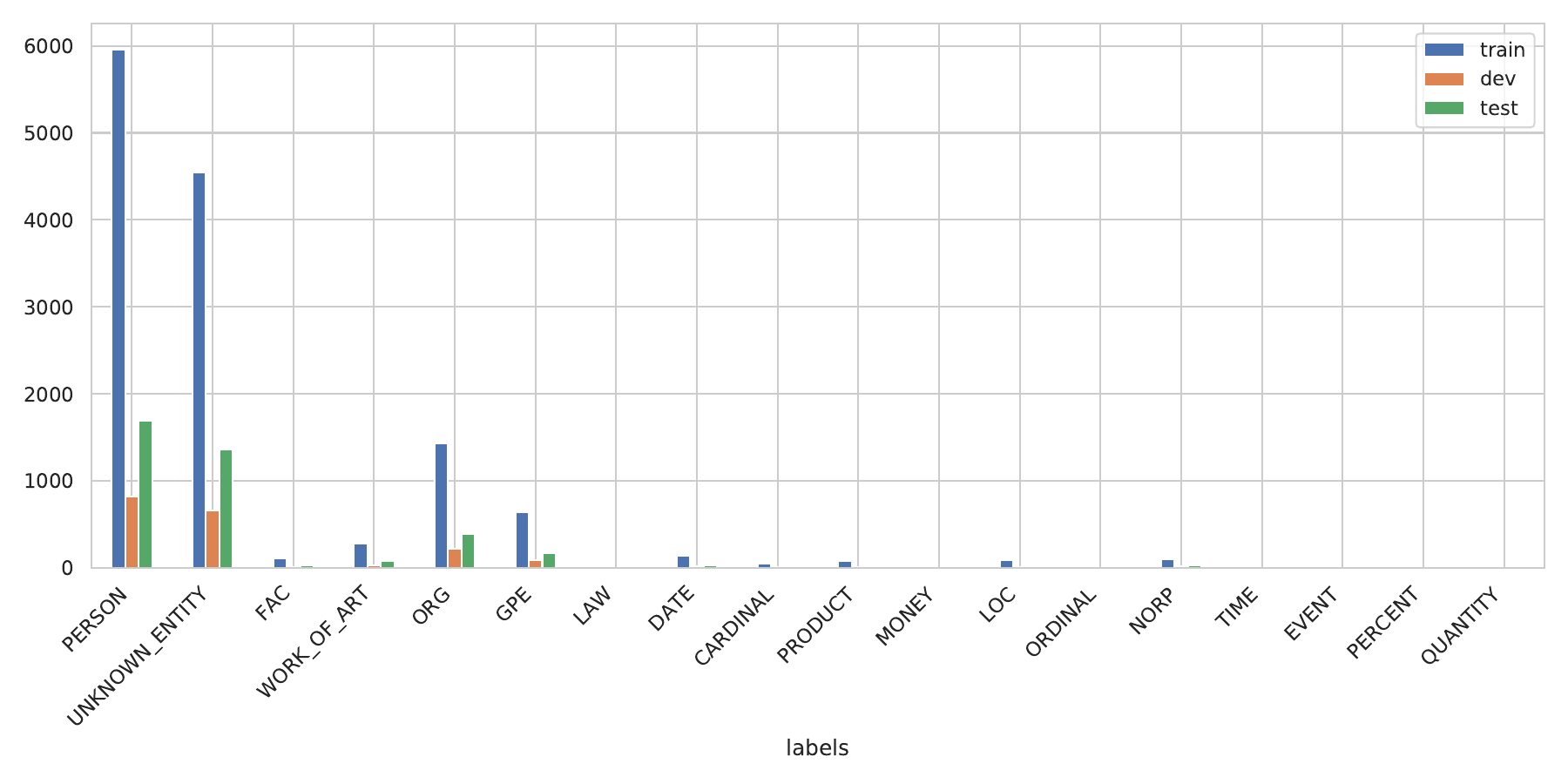}
    \caption{Label distribution for supervised entity type.} 
    \label{fig:distr}
\end{figure*}

The data distribution is shown in \autoref{fig:distr}. As can be seen, it is very skewed.

\subsubsection{Models}
We use two types of models: 1) a \finetuned\ 12 layer 768 dimensional \bertbc\ model; and 2) a popular word convolutional model for sentence classification \cite{DBLP:conf/emnlp/Kim14} using three parallel filters (size 3, 4, and 5) and 300 dimensional Google News Word2Vec representations \cite{DBLP:journals/corr/abs-1301-3781}. 

\textbf{BERT model}: This model is trained for 5 epochs, with Adam optimizer \cite{DBLP:journals/corr/KingmaB14} with a weight decay of \texttt{1.0e-08} and a learning rate of \texttt{1.0e-05}. The sequence max length is kept at 128. We search for two hyperparameters: 1) number of epochs: 3-7, increasing by 1; and 2) learning rate: \texttt{1.0e-05}, \texttt{5.0e-05}, \texttt{1.0e-04}. Accuracy is used as the early stopping metric.

\textbf{WordConv model}: This model is trained for 40 epochs, with Adadelta optimizer \cite{DBLP:journals/corr/abs-1212-5701} with a learning rate of \texttt{1.0e-05}. The sequence max length is again kept at 128. Accuracy is used as the early stopping metric.

\subsection{Results}
The results are provided in \autoref{tab:supervised-etype-results}.

\begin{table}[!t]
\begin{tabular}{|l|l|l|}
\hline
         & acc              & macro\_f1        \\ \hline
\bertbc\     & 63.55 $\pm$ 0.00 & 19.65 $\pm$ 0.04 \\ \hline
WordConv & 58.81 $\pm$ 0.01 & 13.31 $\pm$ 0.02 \\ \hline
\end{tabular}
\caption{Results for supervised entity type selection}
\label{tab:supervised-etype-results}
\end{table}

\end{document}